\documentclass[journal]{IEEEtran}

\usepackage{graphicx}
\usepackage{amsmath,graphicx,bm,threeparttable,indentfirst,cite}
\usepackage{algorithm, algorithmic,booktabs}
\usepackage{color, multirow,graphicx}
\usepackage{multirow,epsfig,fbox,amsfonts,amsmath,multicol,enumitem,color,pifont}
\usepackage{hyperref}
\usepackage{graphicx}

\begin{document}

\title{Auto-Prompting SAM for Weakly Supervised Landslide Extraction}

\author{Jian~Wang,        Xiaokang~Zhang,~\IEEEmembership{Senior Member,~IEEE,}
Xianping~Ma,~\IEEEmembership{Graduate Student Member,~IEEE,}
Weikang~Yu,~\IEEEmembership{Graduate Student Member,~IEEE,}
and~Pedram~Ghamisi,~\IEEEmembership{Senior Member,~IEEE}% <-this % stops a space
\thanks{This work was supported in part by the National Natural Science
Foundation of China under Grant 42371374. \textit{(Corresponding author: Xiaokang Zhang.)}}
\thanks{Jian Wang and Xiaokang Zhang are with the School of Information Science and Engineering, Wuhan University of Science and Technology, Wuhan 430081, China (e-mail: natezhangxk@gmail.com).}% <-this % stops a space
\thanks{Xianping Ma is with the School of Science and Engineering, The Chinese University of Hong Kong, Shenzhen 518172, China (e-mail: xianpingma@link.cuhk.edu.cn).}
\thanks{Weikang Yu is with the Chair of Data Science in Earth Observation, Technical University of Munich (TUM), 80333 Munich, Germany, and also with Helmholtz-Zentrum Dresden-Rossendorf, 09599 Freiberg, Germany (e-mail: w.yu@hzdr.de).}% <-this % stops a space
\thanks{Pedram Ghamisi is with Helmholtz-Zentrum Dresden-Rossendorf, 09599 Freiberg, Germany,
and also with the Lancaster Environment Centre, Lancaster University, LA1 4YR Lancaster, U.K. (e-mail: p.ghamisi@hzdr.de).}}

\markboth{Journal of \LaTeX\ Class Files,~Vol.~13, No.~9, September~2014}%
{Shell \MakeLowercase{\textit{et al.}}: Bare Demo of IEEEtran.cls for Journals}

\maketitle

\begin{abstract}
Weakly supervised landslide extraction aims to identify landslide regions from remote sensing data using models trained with weak labels, particularly image-level labels. However, it is often challenged by the imprecise boundaries of the extracted objects due to the lack of pixel-wise supervision and the properties of landslide objects. To tackle these issues, we propose a simple yet effective method by auto-prompting the Segment Anything Model (SAM), i.e., APSAM. Instead of depending on high-quality class activation maps (CAMs) for pseudo-labeling or fine-tuning SAM, our method directly yields fine-grained segmentation masks from SAM inference through prompt engineering. Specifically, it adaptively generates hybrid prompts from the CAMs obtained by an object localization network. To provide sufficient information for SAM prompting, an adaptive prompt generation (APG) algorithm is designed to fully leverage the visual patterns of CAMs, enabling the efficient generation of pseudo-masks for landslide extraction.
These informative prompts are able to identify the extent of landslide areas (box prompts) and denote the centers of landslide objects (point prompts), guiding SAM in landslide segmentation. Experimental results on high-resolution aerial and satellite datasets demonstrate the effectiveness of our method, achieving improvements of at least 3.0\% in F1 score and 3.69\% in IoU compared to other state-of-the-art methods. The source codes and datasets will be available at \url{https://github.com/zxk688}.
\end{abstract}


\IEEEpeerreviewmaketitle

\section{Introduction}
\IEEEPARstart{L}{andslides}, one of the most common natural hazards worldwide, cause significant economic damage and result in numerous human casualties each year. Accurate recording of landslide characteristics, such as size, location, and spatial extent, is crucial for hazard and risk assessment. Landslide extraction from high-resolution remote sensing imagery has gained increasing attention with advancements in remote sensing technology \cite{9921261}. Recently, deep learning techniques have demonstrated their effectiveness in pixel-wise landslide extraction from high-resolution imagery \cite{zhang2023cross}. However, the high performance of landslide extraction models relies heavily on large volumes of training data. The manual annotation process incurs substantial costs, and both unlabeled data and image-level labeled data are relatively easy to acquire \cite{yue2022optical}. Therefore, recent studies have investigated weakly supervised learning to address this issue, which utilizes image-level labels for object extraction. However, such supervision only indicates the presence of a specific object category without providing information about the object's spectral or spatial characteristics.

Existing weakly supervised methods have leveraged the class activation maps (CAMs) to generate high-quality pseudo-masks \cite{zhou2016learning}. Previous works mainly focus on improving the quality of CAMs to capture more complete and informative regions of desired objects because they directly obtain pseudo-labels through the CAMs.
For example, they attempt to exploit local-global semantic contexts for feature representation or pixel-level spatial contextual relationships based on pixel affinities \cite{zhang2023weakly,qiao2023weakly}. However, it is still challenging to accurately locate and extract the landslide objects under weak supervision due to their irregular shapes, fragmented boundaries, and scattered distribution.  

Recently, the segment Anything Model (SAM) \cite{kirillov2023segment} has demonstrated remarkable zeo-shot segmentation performance in remote sensing \cite{10636322}.
While SAM is powerful and flexible, it relies on precise prompts to guide the segmentation of objects. It cannot automatically identify the desired objects without such guidance, especially true in complex environments. For remote sensing image scenarios, ambiguous or incorrect prompts can lead to unsatisfactory results. To this end, parameter-efficient fine-tuning by introducing additional modules or learnable prompt embeddings has been explored in remote sensing segmentation \cite{10409216}. In weakly supervised learning, SAM has been utilized to refine the object boundaries in pseudo-labels or directly to fine-tune the SAM encoder using the pseudo-labels \cite{10570484, 10310006} after CAM-based methods were exploited to generate initial pseudo-labels. However, they mainly focus on segmenting objects with well-defined edges, such as buildings and urban scenes. Due to the complex background interference and weak object edges in landslide-related images, SAM's capabilities of instance segmentation and boundary refinement significantly degrade. 

% The interactive and prompt-driven design allows for user-directed segmentation of objects in diverse images. understanding of remote-sensing images. Furthermore, we note that the complex background interference and the absence of well-defined object edges in remote sensing image scenarios pose significant challenges to SAM’s segmentation capabilities

% Prompts don't carry global or relational context. For example, SAM cannot inherently understand "segment all cars in the image" unless multiple prompts or extensions are used. While SAM is flexible, it requires user guides for segmentation and cannot automatically identify objects without guidance \cite{}. SAM's outputs depend on the quality and specificity of the prompt. However, ambiguous or incorrect prompts can lead to unsatisfactory results.

To address the abovementioned issues, a weakly supervised landslide extraction method called APSAM based on an auto-prompting algorithm is proposed. Instead of adapting or fine-tuning the SAM encoder for feature extraction using pixel-level labels \cite{10409216}, we utilized the object localization capability of CAM-based methods with image-level supervision, which can guide SAM in producing fine-grained segmentation masks. Specifically, it adaptively explores hybrid prompts from an object localization network with local-global semantics enhancement. These prompts can identify the extent of landslide areas (box prompts) and represent landslide objects (point prompts), guiding SAM in landslide segmentation. To provide sufficient information for SAM, an adaptive prompt generation (APG) algorithm is designed to fully exploit the visual patterns of CAMs. As a result, reliable pseudo-masks for landslide extraction can be effectively obtained by leveraging these prompts.

% I wish you the best of success.

 
% \hfill September 17, 2014
\begin{figure}[h]
    \centering
    \includegraphics[width=1\linewidth]{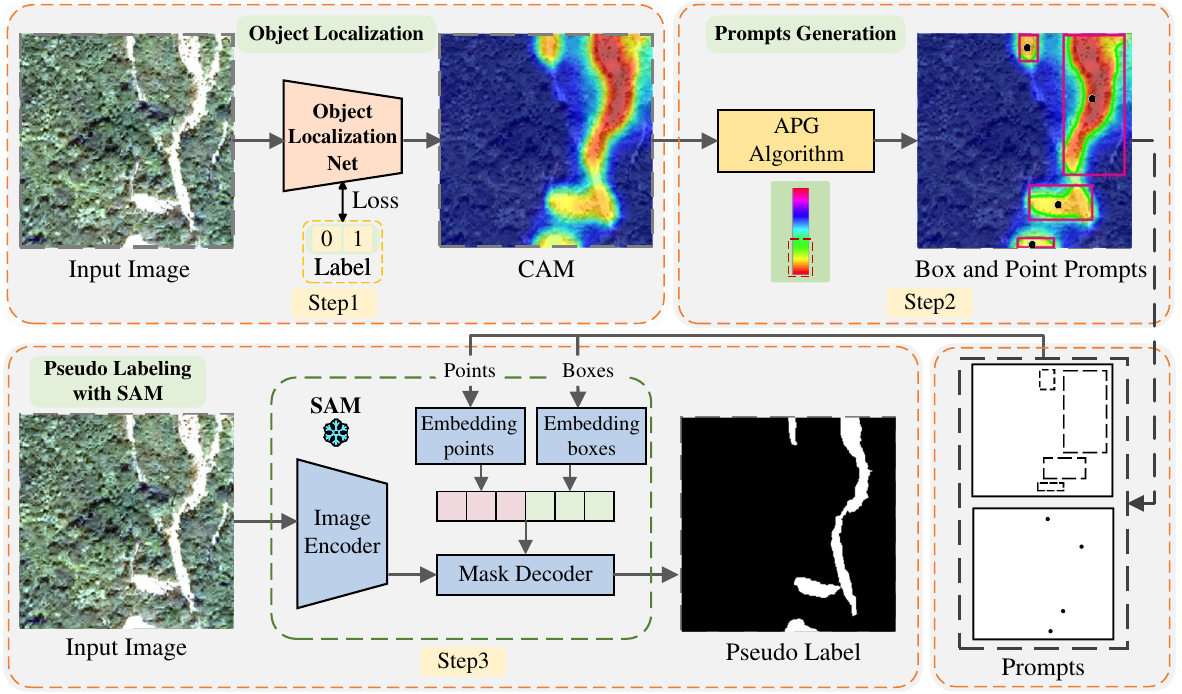}
    \caption{Overview of the proposed APSAM for landslide extraction.}
    \label{fig:fk}
\end{figure}

\section{Methodology}
This section provides a detailed description of APSAM, as illustrated in Fig.~\ref{fig:fk}. It consists of three stages. First, an object-localization network is utilized to generate CAMs. Then,  an adaptive prompt generation algorithm is applied to automatically create informative point and box prompts. After that, the pre-trained SAM is employed to extract landslides and generate pseudo-masks based on the generated prompts.

\subsection{Object Localization}
To coarsely locate the landslides and obtain initial seed regions, an object localization network \cite{zhou2016learning} is trained and utilized in the proposed approach. Given the training data \(\{\boldsymbol{X_i}, \boldsymbol{Y_i}\}_{i=1}^{N}\) with \(N\) samples, where \(\boldsymbol X\) is the input image and \(\boldsymbol{Y}\in \{0, 1\}\) represents the image-level label. The coarse $\boldsymbol{CAM}$ can be calculated by:
\begin{equation}
\boldsymbol{CAM}_{i, j}(x,y)=\frac{\theta_{j}^{\mathrm{T}} \boldsymbol{F}_{i}(x,y)}{\max _{x,y} \theta_{j}^{\mathrm{T}} \boldsymbol{F}_{i}(x,y)} 
\label{eq1}
\end{equation}
where $(x, y)$  indexes the spatial location in the input image, $\theta_{j}$ denotes the weights of the \(j\)th category and  \(\boldsymbol{F_i} \in \mathbb{R}^{C\times H\times W}\) represents the feature map embedded by the backbone. Notably, although the activation map can focus on the most discriminative parts of the feature map, it typically only provides coarse object localization without precise boundaries of objects.
% By analyzing the structural similarity between the initial CAM and each pixel's structure, it can achieve initial object localization results.
% Specifically, the paper proposes a structure-aware object locating module~(SOL) that replaces the traditional CAM method for extracting seed regions \cite{zhou2016learning}. This module captures more complete seed regions for each class by analyzing the structural similarity between the initial CAM and each pixel's structure. Subsequently, the paper employs a Global Anchor Aggregation~(GAA) module and uses a cross-attention mechanism to inject global semantic contexts into the representation of features, thereby achieving more accurate object localization.

% \subsubsection{Generating Coarse Activation Maps by CAM}

% Here, we provide a brief description of the simple CAM method for generating heatmaps.
% For specific and refined methods of generating heatmaps, please refer to the paper LGAGNet\cite{zhang2023weakly}.

% \subsubsection{structure-aware object locating module}
% In the SOL module, the spatial structure is emphasized by modeling semantic affinities between pixels in the feature map as follows:

\subsection{Adaptive Prompts Generation}
Normally, the discriminative object regions can be obtained by applying a hard threshold to the heatmap of CAM.
However, inaccurate heatmaps with incomplete object localizations could mislead the box prompts in weakly supervised learning. Furthermore, the inferences caused by objects such as trees, rocks, and buildings often lead to non-uniform distributions of pixel prompts. As a result, biased segmentation results could be obtained. To improve the robustness of the prompting, the proposed approach combines point- and box-level prompts through the adaptive prompting algorithm.
% For box prompts, in theory, if the bounding box accurately covers the region, SAM can produce precise segmentation areas. However, in practical weakly supervised learning algorithms, it is often challenging to generate accurate heatmaps, which in turn leads to less accurate boxes. For some images, the heatmap may only cover a small portion of the object, making the use of point prompts more effective.  And then for point prompts, satisfactory results can often be achieved if the pixels inside the object to be segmented are uniform. However, in areas like landslides, various objects such as trees, rocks, and buildings may interfere, leading to non-uniform pixel distributions. If a point is placed in an area with uneven pixels, the segmentation results can be biased. In contrast, bounding boxes do not encounter this issue. For a detailed comparison, please refer to the experimental section. Therefore, our approach utilizes both points and boxes as prompts, resulting in commendable outcomes.

% In this section, we will introduce how to generate point and box prompts for each input image from the heatmap. 
The implementation of the algorithm is illustrated as follows. First, the binarization on the generated heatmap is performed to extract the regions of interest. For each pixel, if its grayscale value exceeds the threshold \(T\in[0,~255]\), that pixel is marked as foreground in the binary image. Otherwise, it is marked as background. Let \(H(x,y)\) represent the grayscale value of the pixel \((x,y)\) in the heatmap \(\boldsymbol{CAM}\in\mathbb{R}^{H\times W}\). The binarization can be expressed as follows: 
\begin{equation}
\boldsymbol{B}(x, y)=\left\{\begin{array}{ll} 1, & \text { if } H(x, y)>T \\ 0, & \text { otherwise } \end{array}\right.
\label{eq2}
\end{equation}
where \(\boldsymbol{B}(x, y)\) is the resulting binary image. Once the specific binary regions are obtained, the contours in the binary image can be detected. After that, a list of contours along with hierarchy information can be obtained:
\begin{equation}
\boldsymbol{C} = \{c_1, c_2, ...c_K\} = findContours(\boldsymbol{B}(x,y))
\label{eq3}
\end{equation}
where $K$ is the number of contours and \(c_k\) denotes the \(k\)th contour in the contour set \(\boldsymbol{C}\). Then for each detected contour, the coordinates of its bounding box can be  calculated as follows:
\begin{equation}
\boldsymbol{b}_k = (x_0,~y_0,~w,~h) = boundingRect(c_k)
\label{eq4}
\end{equation}
where \(\boldsymbol{b}_k\) denotes bounding box of \(c_k\),  \((x_0,~y_0)\) is the top-left corner of the box, and \(w\) and \(h\) are its width and height.
On this basis, we compute the centroid of each contour as the point prompt \(\boldsymbol{p}_k\). Feature moments are used to calculate various characteristics:
\begin{equation}
% m = CV2.moments(c_k)
m_{p,q} = \sum_{x}\sum_{y} x^p y^q c_k(x,y)
\label{eq5}
\end{equation}
where \(m\) is the moments, which include multiple moment values. Then the centroid coordinates \((c_x,c_y)\) can be computed using
\begin{equation}
\boldsymbol{p}_k = \begin{cases} \left(\frac{m_{1,0}}{m_{0,0}}, \frac{m_{0,1}}{m_{0,0}}\right), & \text{if } m_{0,0} > 0 \\ \text{undefined}, & \text{otherwise} \end{cases}
% \boldsymbol{p}_k = (c_x,c_y),~c_{x}=\frac{m_{1,0}}{m_{0,0}},~c_{y}=\frac{m_{0,1}}{m_{0,0}}
\label{eq6}
\end{equation}
where \(m_{1,0}\) and \(m_{0,1}\) denote the first moment of the contour, representing the accumulated value of $x$ and $y$ separately, and \(m_{0,0}\) denotes the zero-order moment of the contour, representing the area of the contour. 
Finally, the generated point and box prompts for SAM are as follows:
\begin{equation}
\boldsymbol{P_i} = \{\boldsymbol{p}_1,~\boldsymbol{p}_2,...,\boldsymbol{p}_K\},
\label{eq7}
\end{equation}
\begin{equation}
\begin{aligned}
\boldsymbol{B_i} = \{\boldsymbol{b}_1,~\boldsymbol{b}_2,...,\boldsymbol{b}_K\}
\label{eq8}
\end{aligned}
\end{equation}
where \(\boldsymbol{P_i}\) and \(\boldsymbol{B_i}\) represent all the prompts generated from an image \(\boldsymbol{X_i}\) with \(K\) contours. 

\subsection{Pseudo Labeling with SAM}
% The Segment Anything Model (SAM) \cite{kirillov2023segment}, developed by Meta AI Research as part of the Segment Anything (SA) project, is a cutting-edge image segmentation model that demonstrates impressive zero-shot performance and robust generalization across various application scenarios~ref{}. 
SAM incorporates a Vision Transformer-based image encoder, a lightweight mask decoder, and a versatile prompt encoder that can accept inputs such as points, bounding boxes, masks, and text \cite{kirillov2023segment}. The variety of prompt methods allows users to focus on different scenes and objects, providing SAM with strong flexibility for diverse applications. 
% Drawing inspiration from the success of Large Language Models (LLMs) in natural language processing (NLP), SAM's image encoder follows the Vision Transformer architecture and offers three model sizes: ViT-B, ViT-H, and ViT-L. The MAE~\cite{he2022masked} method is utilized to pretrain the Vision Transformer, enabling it to process high-resolution input images. SAM's powerful zero-shot capabilities are achieved through prompt engineering, which includes two types of prompts: sparse prompts (points, boxes, and text) and dense prompts (masks). Positional encoding~\cite{tancik2020fourier} is used to extract embeddings for the sparse prompts, while CLIP~\cite{radford2021learning} is responsible for text prompts. Dense prompts are embedded via a convolutional module and then fused elementwise with the image embeddings. 
% In the proposed approach,  the automatically generated point and box prompts are combined and fed into the SAM to generate more reliable pseudo-labels. The process can be described as follows:
In the proposed approach,  the automatically generated point and box prompts are combined and fed into the SAM. After that, SAM encodes the point prompts by embedding their spatial coordinates and labels (with all labels being positive) and separately encodes the box prompts. These embeddings are then combined into a unified sparse embedding, which is input into SAM's mask decoder to generate accurate pseudo-labels. The process can be represented as follows:
\begin{equation}
\hat{\boldsymbol{Y}_i} = \boldsymbol{\mathcal{D}_{mask}}(\boldsymbol{\mathcal{E}_{img}}(\boldsymbol{X_i}),~\boldsymbol{\mathcal{E}_{prompt}}(\left[\boldsymbol{P_i},\boldsymbol{B_i}\right]))
\label{eq9}
\end{equation}
where \(\boldsymbol{\mathcal{E}_{img}}\) denotes SAM's image encoder, \(\boldsymbol{\mathcal{E}_{prompt}}\) denotes SAM's prompt encoder, and \(\boldsymbol{\mathcal{D}_{mask}}\) represents SAM's mask decoder. Moreover, \(\hat{\boldsymbol{Y}_i}\) is the pseudo label generated by SAM.

\begin{algorithm}[h]
		\caption{{APSAM.}}
	\begin{algorithmic}[1]
		\label{alg1}
		\STATE{\textbf{Input:} \(\{\boldsymbol{X_i}, \boldsymbol{Y_i}\}_{i=1}^{N}\): training data; $\boldsymbol{CAM_i}$ : heatmap of sample \(i\); $SAM$: Segment Anything Model.}
		{\FORALL{\(\{\boldsymbol{X_i}, \boldsymbol{Y_i}\}_{i=1}^{N}\)}

                {\IF{\(\boldsymbol{Y_i} == 0\)}
                \STATE \textbf{Continue}
                \ENDIF}
                \STATE{\(\boldsymbol{B}(x, y)\leftarrow \boldsymbol{CAM_i}\)}
                \STATE{\(\boldsymbol{C} = \{c_1, c_2, ...,c_K\} \leftarrow \boldsymbol{B}(x,y)\)}
                {\FORALL{\(c_k\in\boldsymbol{C}\)}
                \STATE{\(\boldsymbol{b}_k \leftarrow c_k\)}
                \STATE{\(\boldsymbol{p}_k \leftarrow m_{p,q} \leftarrow c_k\)}
                \ENDFOR}
                \STATE{\(\boldsymbol{P_i} \leftarrow \{\boldsymbol{p}_1,~\boldsymbol{p}_2,...,\boldsymbol{p}_K\}\)}
                \STATE{\(\boldsymbol{B_i} \leftarrow \{\boldsymbol{b}_1,~\boldsymbol{b}_2,...,\boldsymbol{b}_K\}\)}
                \STATE{\(\hat{\boldsymbol{Y}_i} \leftarrow {SAM}(\boldsymbol{X_i},~\boldsymbol{P_i},~\boldsymbol{B_i} )\) }
			\ENDFOR	}
	\end{algorithmic}
\end{algorithm}

\subsection{Training and Inference}
The object localization module is trained with image-level labels to generate CAMs. Apart from this, no additional training is needed in the proposed APSAM. Prompts are generated automatically, and SAM uses them for inference to implement the segmentation. The procedure for generating pseudo-labels is detailed in Algorithm~\ref{alg1}. Finally, the pseudo-labels are used to retrain a semantic segmentation model for test data predictions.

% to obtain more accurate labels. The specific results can be found in the experimental section. 

%\IEEEpubidadjcol
\begin{figure}[h]
    \centering
    \includegraphics[width=\linewidth]{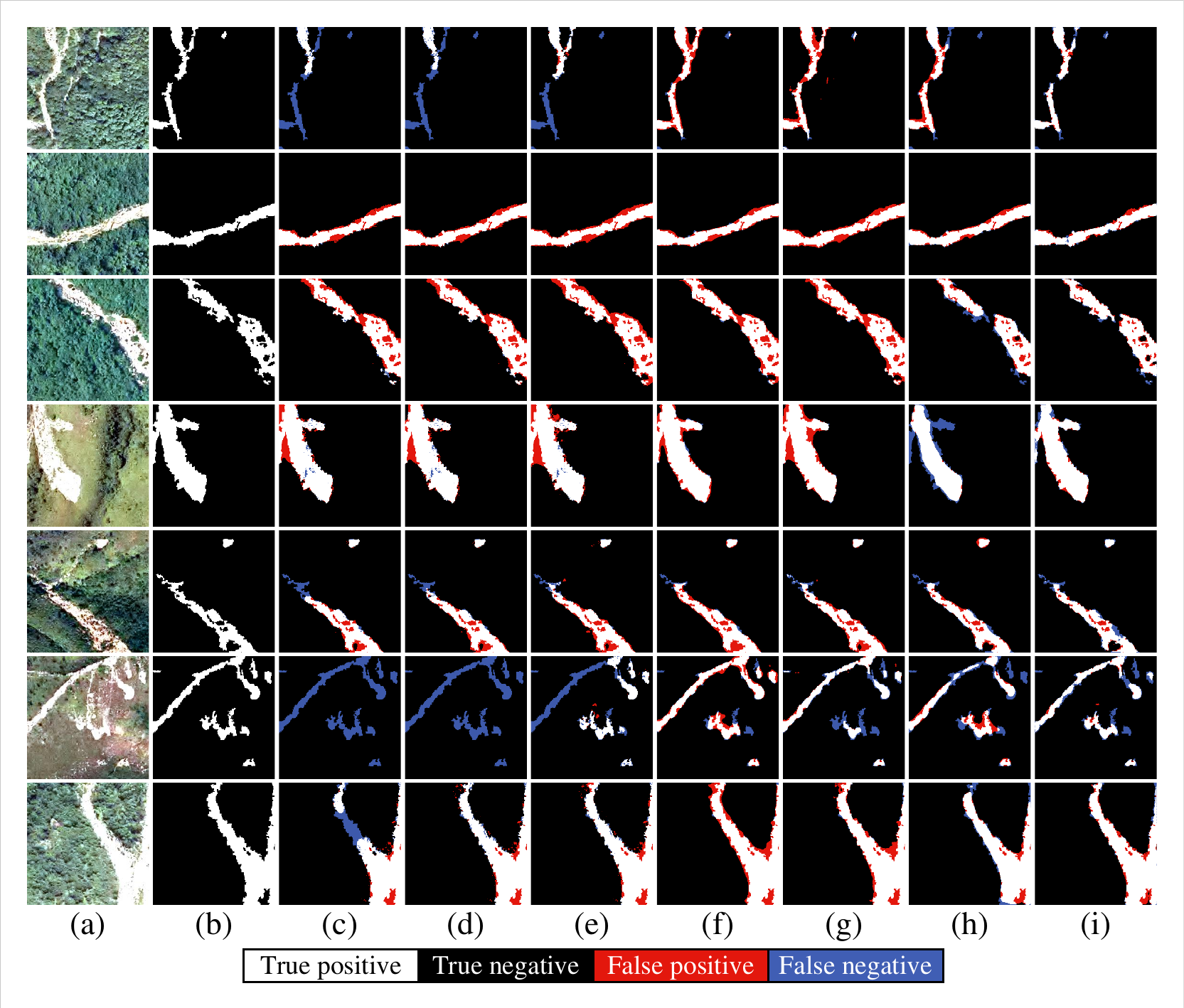}
    \caption{ Landslide extraction results on the Hong Kong dataset. (a) Original
 image. (b) Ground truth. The results were obtained by (c) CDA, (d) CPN,
 (e) CSE, (f) FlipCAM, (g) S2C, (h) LGAGNet, and (i) Our Method. The
 image size is 256 × 256 pixels.}
    \label{fig:HK}
\end{figure}

\begin{figure}[h]
    \centering
    \includegraphics[width=\linewidth]{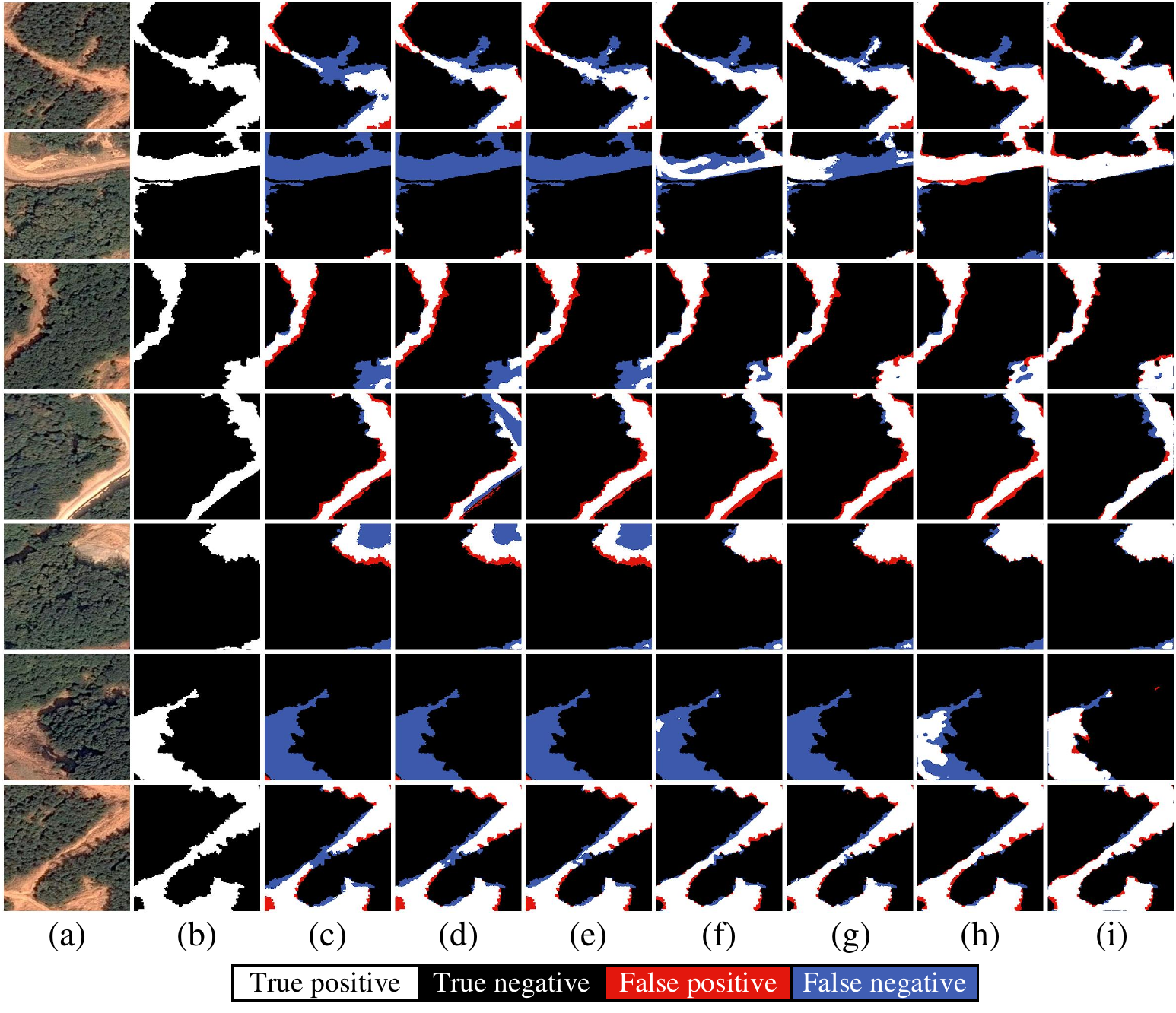}
    \caption{ Landslide extraction results on the Turkey dataset. (a) Original
 image. (b) Ground truth. The results were obtained by (c) CDA, (d) CPN,
 (e) CSE, (f) FlipCAM, (g) S2C, (h) LGAGNet, and (i) Our Method. The
 image size is 256 × 256 pixels.}
    \label{fig:TK}
\end{figure}

\section{Experimental Results}
\subsection{Data Description}
Two high-resolution datasets, Hong Kong and Turkey, are used to evaluate the proposed method \cite{zhang2023cross}. The Hong Kong dataset contains images captured in 2014, with a spatial resolution of 0.5 m, consisting of aerial images. The TurKey dataset contains images captured in 2017, with a spatial resolution of 0.59 m, consisting of Google Earth images, covering landslides caused by heavy rain in parts of Turkey. Both datasets have their images cropped to a size of 256 $\times$ 256, with a 40\% overlap between adjacent images. The Hong Kong dataset uses 630 images for training with image-level labels and tests with pixel-level labels. The TurKey dataset uses 1590 images for training with image-level labels and tests with pixel-level labels.

\begin{table}
\parbox{\linewidth}{%
\centering
\caption{QUANTITATIVE COMPARISON (\%) OF DIFFERENT METHODS FOR  weakly supervised LANDSLIDE EXTRACTION.}
\setlength{\tabcolsep}{4pt}
\begin{tabular}{ccccccc} 
\hline
Dataset                    & Method     & OA    & Precision & Recall & F1    & IoU    \\ 
\hline
\multirow{7}{*}{Hong Kong} & CDA \cite{su2021context}        & 97.89 & \textbf{80.51}     & 64.78  & 71.79 & 56.00  \\
                           & CPN \cite{zhang2021complementary}       & 97.92 & 76.82     & 71.55  & 74.09 & 58.85  \\
                           & CSE \cite{kweon2021unlocking}       & 97.77 & 69.89     & 81.21  & 75.13 & 60.16  \\
                           & FlipCAM \cite{10416708}       & 96.46 & 55.36     & 75.65  & 63.93 & 46.99  \\
                           & S2C \cite{kweon2024sam}       & 95.85 & 50.00     & 68.74  & 57.89 & 40.73  \\
                           & LGAGNet \cite{zhang2023weakly}    & 98.02 & 71.37     & \textbf{87.52}  & 78.62 & 64.78  \\
                           & Ours       & \textbf{98.24} & 77.90   & 80.42 & \textbf{79.14} & \textbf{65.48} \\
\hline
\multirow{7}{*}{Turkey}    & CDA \cite{su2021context}       & 97.72 & 76.04     & 48.35  & 59.12 & 41.96  \\
                           & CPN \cite{zhang2021complementary}       & 97.87 & 73.54     & 58.57  & 65.21 & 48.38  \\
                           & CSE \cite{kweon2021unlocking}       & 97.75 & 73.04     & 53.89  & 62.02 & 44.95  \\
                           & FlipCAM \cite{10416708}       & 97.87 & 76.28     & 54.53  & 63.59 & 46.62  \\
                           & S2C \cite{kweon2024sam}       & 98.09 & 75.08     & 65.91  & 70.20  & 54.08  \\
                           & LGAGNet \cite{zhang2023weakly}    & 98.18 & \textbf{78.25}     & 64.48  & 70.70  & 54.68  \\
                           & Ours       &\textbf{98.20} & 73.31 & \textbf{74.12} & \textbf{73.71} & \textbf{58.37} \\
\hline
\label{tb1}
\end{tabular}
}
\end{table}

\subsection{Experimental setup}
In the object localization section, original CAM \cite{zhou2016learning} is employed to generate heatmaps. Furthermore, the SGD optimizer is chosen with a weight decay of 0.0001 and a momentum of 0.9 to train the network. The learning rate was initially set to 0.1 with a step-wise decay scheduler applied. In the prompts generation section, the threshold \(T\) is set to 120. The third step utilizes SAM's powerful prompt segmentation capabilities to generate pseudo-labels. Finally, to ensure fairness in the experiment, pseudo-labels generated by various methods are trained on ResUNet to obtain the final results. The proposed approach was compared with several state-of-the-art weakly supervised methods, {context decoupling augmentation~(CDA)~\cite{su2021context}, complementary patch network~(CPN)~\cite{zhang2021complementary}, class-specific erasing method~(CSE)~\cite{kweon2021unlocking}, feature-level flipping CAM~(FilpCAM)~\cite{10416708}, from-SAM-to-CAMs~(S2C)~\cite{kweon2024sam}, local–global anchor guidance network~(LGAGNet) \cite{zhang2023weakly}.} For quantitative evaluation, five commonly used metrics were employed: overall accuracy (OA), precision, recall, F1-score, and IoU.

\subsection{Results}
As shown in Fig.~\ref{fig:HK}, the extraction results of various irregular-shaped landslides in the Hong Kong dataset are presented. It can be observed that due to the different learning abilities of various weakly-supervised methods for landslides of different shapes, some methods struggle to detect and generate reliable landslide regions, such as CDA and CPN, where large areas of undetected regions exist in some areas. In comparison to other methods, our method achieves more precise boundaries and performs well even in cases of hollow areas, which are difficult to predict. From the Tabel~\ref{tb1}, we can see that while CDA and LGAGNet excel in Precision and Recall, they fall short in F1 score and IoU, especially CDA, which has a relatively low IoU. This means that although these methods result in fewer misclassified landslide regions, they perform poorly in overall segmentation accuracy and the overlap of target regions. Our network is able to better preserve detailed information on the landslide objects and achieves the highest OA, F1-Score, and IoU values.

As shown in Fig.~\ref{fig:TK}, the extraction results of various irregularly shaped landslides from the Turkey dataset are presented. It can be observed that due to the varying learning capabilities of different weakly supervised methods for different shaped landslides, some methods, such as CDA, CPN, and CSE, exhibit significant areas of missed detections in certain regions. In contrast, our method achieves more accurate and consistent boundaries with fewer misclassifications. As indicated in Table~\ref{tb1}, although LGAGNet demonstrates competitive performance in overall Precision, reaching 78.25\%, indicating its precision in landslide detection, our method achieves a higher Recall, resulting in fewer missed detections. Compared to other methods, our network achieves the highest values for OA, Recall, F1-score, and IoU, with scores of 98.20\%, 74.12\%, 73.71\%, and 58.31\%, respectively. The IoU has a significant gain of 3.69\% to 16.41\%, proving that our method not only accurately detects landslides but also better preserves detailed information and effectively handles complex shapes.
\begin{figure}[h]
    \centering
    \includegraphics[width=\linewidth]{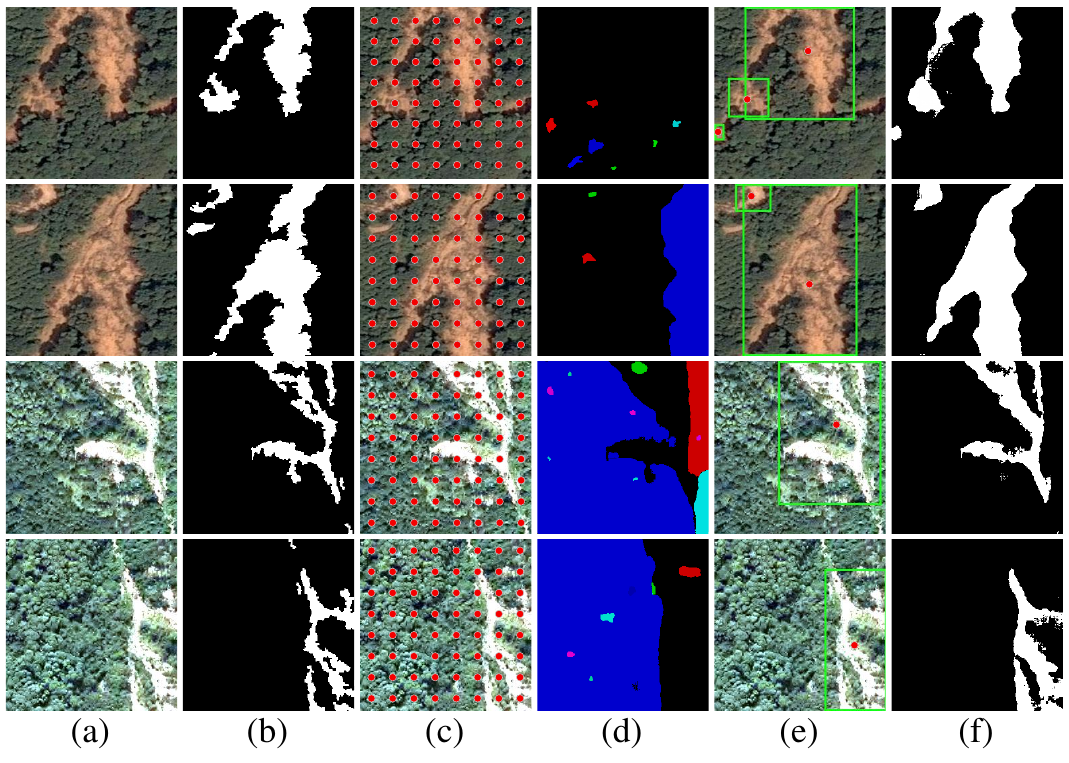}
    \caption{ Comparison of prompting results obtained by SAM and APSAM. (a) Original
 image. (b) Ground truth. (c) SAM's point prompts. (d) Overlay result of SAM segmentation by color mask. (e) APSAM's box and point prompts. (f) Result of APSAM segmentation.}
    \label{fig:cmp2}
\end{figure}

\subsection{Analysis}
We conducted two ablation experiments on the Hong Kong and Turkey datasets to evaluate the effectiveness of different prompts in APSAM, as well as the segmentation results obtained directly using SAM compared to those generated by APSAM.
From Fig.~\ref{fig:cmp2}, it can be seen that using SAM for full-image segmentation produces prompts uniformly across the entire image. The number of points that can be generated along one edge is adjustable and set to N; in this case, we conducted the experiment with $N=8$. The experimental results indicate that full-image segmentation using SAM can only produce contour information and does not distinguish specific semantic details. As shown in the colored results Fig.~\ref{fig:cmp2}~(d), SAM tends to select the contour areas in the image for segmentation. In contrast, APSAM generates more accurate point and box prompts through a heatmap, resulting in better segmentation outcomes.

\begin{table}
\parbox{\linewidth}{%
\centering
\caption{QUANTITATIVE COMPARISON (\%) OF DIFFERENT COMBINATIONS OF PROMPTS WITH APSAM.}
\setlength{\tabcolsep}{4pt}
\begin{tabular}{ccccccc}
\hline
Dataset   & Prompts       & OA    & Precision & Recall & F1    & IoU   \\
\hline
\multirow{3}{*}{Turkey}    & Point~\(+\)~Box & 97.66 & 64.00        & \textbf{71.03}  & \textbf{67.33} & \textbf{50.75} \\
          & Box        & \textbf{97.71} & \textbf{66.35}     & 66.20   & 66.27 & 49.56 \\
          & Point       & 97.20  & 56.69     & 61.42  & 58.96 & 41.80  \\
\hline
\multirow{3}{*}{Hong Kong} & Point~\(+\)~Box & 96.61 & 56.57     & \textbf{78.54}  & \textbf{65.77} & \textbf{49.00}    \\
          & Box        & \textbf{96.70}  & \textbf{57.92}     & 74.91  & 65.33 & 48.51 \\
          & Point       & 95.51 & 45.36     & 40.74  & 42.92 & 27.33  \\

\hline
\label{tb2}
\end{tabular}
}
\end{table}

\begin{figure}[!h]
    \centering
    \includegraphics[width=0.8\linewidth]{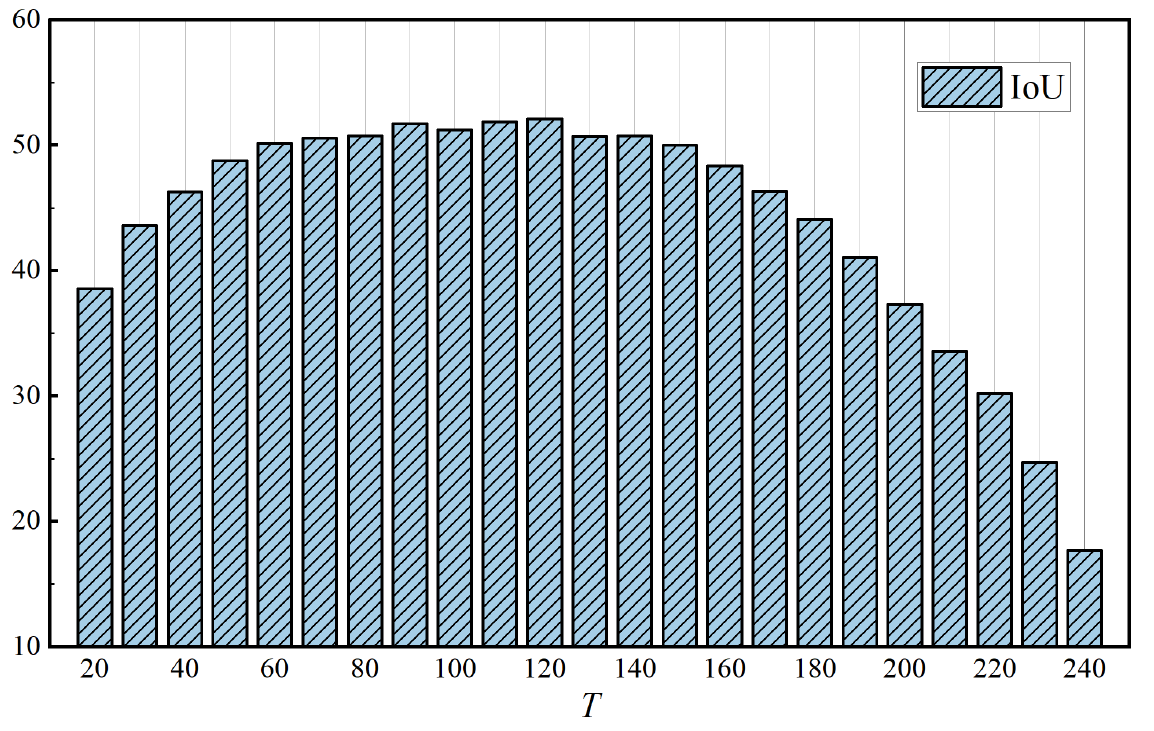}
    \caption{Parameter analysis of the threshold \(T\).}
    \label{pa:L}
\end{figure}

Table~\ref{tb2} illustrates different prompt comparison experiments conducted on the Turkey and Hong Kong datasets. We can see that, for point and box prompts, this method achieved an accuracy of 97.66\%, indicating that it effectively utilized both points and boxes in the segmentation task. Notably, the precision 71.03\%, recall 67.33\%, and F1-score 50.75\% also reflect a balanced performance, particularly in terms of precision. When using only box prompts, the accuracy slightly increased to 97.71\%, suggesting that box-based segmentation may provide a marginal advantage in certain scenarios. However, other metrics showed a decrease compared to the combined method, indicating a reduction in overall effectiveness. When segmentation is performed using only point prompts, all metrics decline. This suggests that while point-based methods can be effective, they may miss the precise boundaries captured by box prompts.

Furthermore, we analyzed the threshold \(T\) in the binarization function of Equation~\ref{eq2}. As shown in Fig.~\ref{pa:L}, it can be seen that the IoU values are stable with the values of generally around $50\%$ when the grayscale threshold \(L\) is between 50 and 170. The optimal value is approximately 120, indicating that this threshold can achieve good results without requiring careful design.

\section{Conclusion}
This study presents a novel weakly supervised landslide extraction network based on the Segment Anything Model (SAM), named APSAM. Instead of focusing on generating precise CAMs for pixel-wise pseudo-labeling, our approach attempts to leverage an adaptive prompts generation (APG) algorithm to obtain accurate prompts from CAMs, which are then fed into the SAM model to produce precise pseudo-labels. The generated prompts, including points and boxes, have adaptive extents and distributions, enabling them to handle the characteristics of landslide objects, such as irregular shapes, fragmented boundaries, and uncertain locations. 
% Furthermore, it has low requirements for the quality of CAMs and does not require direct fine-tuning of SAM, making it easy to deploy. 
Experiments on two high-resolution remote sensing datasets show that APSAM outperforms other state-of-the-art weakly supervised methods. Future work will focus on end-to-end training and enhancing the prompt generation algorithm by incorporating deep learning modules for improved accuracy and flexibility.

\ifCLASSOPTIONcaptionsoff
  \newpage
\fi
\small
\bibliographystyle{IEEEtran}
\bibliography{refs}

\end{document}